\documentclass[]{fairmeta}

\usepackage[utf8]{inputenc} 
\usepackage[T1]{fontenc}    
\usepackage{url}            
\usepackage{booktabs}       
\usepackage{amsfonts}       
\usepackage{nicefrac}       
\usepackage{microtype}      
\usepackage{xcolor}         
\usepackage{enumitem}
\usepackage{comment}
\usepackage{csquotes}

\usepackage{amsmath}
\usepackage{algorithm}
\usepackage{algpseudocode}
\usepackage{graphicx}
\usepackage{amsthm}
\usepackage{amssymb}
\usepackage{thmtools}
\usepackage{thm-restate}
\usepackage{wrapfig}

\usepackage{enumitem} 

\definecolor{highlightcolor}{RGB}{255,255,0}
\newcommand{\highlight}[1]{%
  \setlength{\fboxsep}{0pt}%
  \colorbox{highlightcolor}{$#1$}%
}

\usepackage[colorinlistoftodos,bordercolor=orange,backgroundcolor=orange!20,linecolor=orange,textsize=scriptsize]{todonotes}

\title{Why Gradients Rapidly Increase \\ Near the End of Training}

\author[1]{%
  Aaron Defazio
}

\affiliation[1]{FAIR at Meta}

\abstract{During long-duration Large Language Model (LLM) training runs the gradient norm increases rapidly near the end of training. In this short note, we show that this increase is due to an unintended interaction between weight decay, normalization layers, and the learning rate schedule. We propose a simple correction that fixes this behavior while also resulting in lower loss values throughout training.}

\date{\today}
\correspondence{\email{adefazio@meta.com}}

\begin{document}

\maketitle

\section{Introduction}

\begin{figure}[b!]
  \centering
  \includegraphics[width=\linewidth]{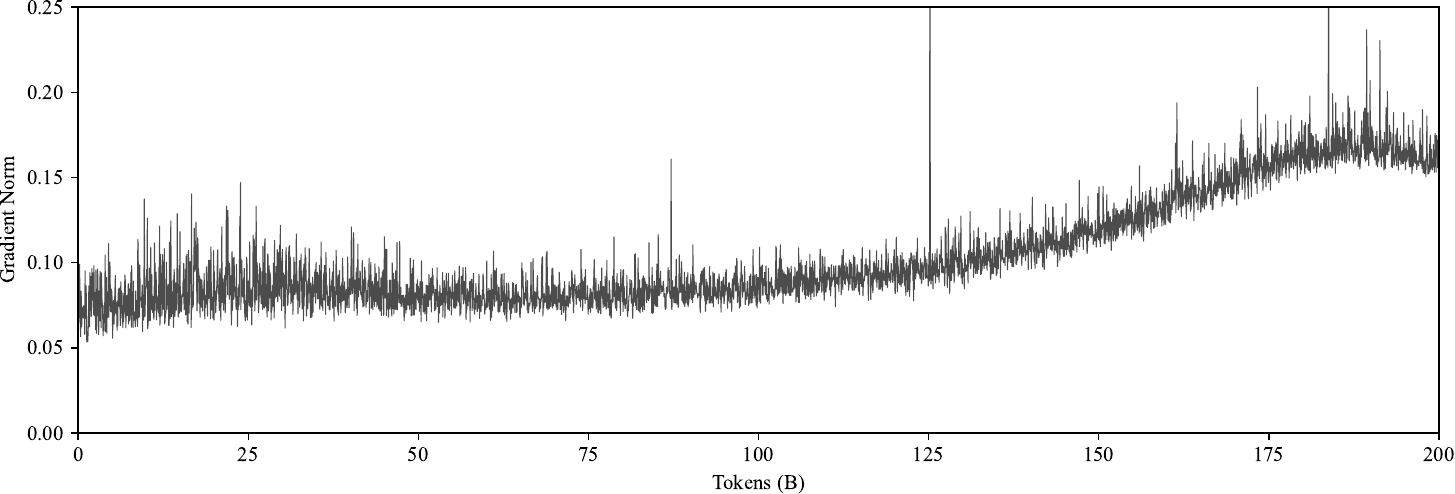}
  \caption{A 120M parameter LLM training run on FineWeb-Edu, showing the behavior where the gradient norm more than doubles towards the end of training.}
  \label{fig:gradient-uptick}
\end{figure}

The health of large-scale training runs is typically monitored by logging key quantities such as the norm of weights, gradients, and activations. The behavior observed in practice of these quantities, their magnitude, and shift over time, is still not well understood theoretically. In this work, we focus on one particular unusual behavior that arises only in longer-duration training runs, and not for all machine learning models, an uptick in the norm of the gradient towards the end of training (Figure~\ref{fig:gradient-uptick}). As far as we are aware, this behavior is currently unexplained in the existing literature.

We show that this behavior is due to an interaction between weight decay and learning-rate schedules that occurs only for weight layers that are affected by normalization layers such as LayerNorm \citep{ba2016layer} or BatchNorm \citep{ioffe2015batch}. We present the following results:

\begin{enumerate}
    \item We utilize an existing theoretical framework by \citet{van2017l2} to show that weight decay controls the ratio of the gradient norm to the weight norm during training.
    \item We show that a consequence of this result is that changing the learning rate over time also changes the gradient-norm-to-weight-norm ratios, resulting in undesirable increases in the gradient norm.
    \item We propose a theory-motivated fix to weight decay that eliminates this gradient norm behavior. We show that corrected versions of Adam (AdamC) and SGD (SGDC) improve over the base optimizers.
\end{enumerate}


\section{Related Work}
The effect of weight decay on the behavior of layers during the course of optimization is the central study of our work. Several existing publications touch on this topic. In ``Three Mechanisms of Weight Decay Regularization'', \citet{zhang2018three} argue that weight decay affects the optimization process through:
\begin{enumerate}
    \item \label{itm:lr} Increasing the effective learning rate
    \item \label{itm:reg} Regularizing the input-output Jacobian when K-FAC is used.
    \item \label{itm:2ndorder} Reducing the damping coefficient for second-order optimizers such as K-FAC
\end{enumerate}
Weight decay accomplishes (\ref{itm:lr}) through its damping effect on weight norms during optimization, as first explored by \citet{van2017l2}. As we show in Section~\ref{sec:dynamics}, this only tells part of the story, as per-layer weights are guided toward an equilibrium with improved cross-layer network conditioning, not just a globally increased learning rate. Points (\ref{itm:reg}) and (\ref{itm:2ndorder}) describe previously unexplored interactions between K-FAC and weight decay, but as we focus on SGD and Adam, they are outside the scope of this work.

\citet{xie2023on} describe the phenomenon of increasing gradient norms that arises due to the use of weight decay. Their analysis proposes a convergence rate bound that shows a dependence on the weight decay constant, but their theory doesn't provide a concrete explanation for why gradient norms increase over time. They also propose to apply a schedule to the weight decay to compensate for this. In contrast, our analysis precisely describes why gradient norms increase and is constructive enough to allow us to propose a corrective term that exactly counteracts the gradient norm increase.

\citet{loshchilov2018decoupled} introduced the AdamW variant of the Adam method that decouples the weight decay application from the rest of the update. This method has been adopted by the community due to its overwhelmingly better performance in practice. Recent theoretical justifications for the effectiveness of this modification has centered around it's relation to a partial-proximal step \citep{zhuang2022understanding}. We provide an alternative explanation in Section~\ref{sec:why-adamw}.

\citet{kosson2024rotationalequilibriumweightdecay} view weight decay through the lens of maintaining a rotational equilibrium. Through this framework, a link between the dynamics of layer/neuron learning rates and weight decay is established. The conclusions we draw in our work use a different analysis technique, but we come to many of the same conclusions at a high level regarding the effect of weight decay on learning. They also introduce a \emph{constrained dynamics} optimization wrapper, which explicitly constrains the behavior of normalized layers, a complementary approach to the small modification to weight decay that we suggest in this work.

\citet{dangelo2024needweightdecaymodern} argue that weight decay's primary effect is as a loss stabilization mechanism during multi-pass training runs. For single pass LLM training, they argue that weight decay results in larger effective learning rates and prevents blowup when training in bfloat16. 

Earlier literature on weight decay and training dynamics focuses on problems where overfitting is a major concern, and label noise is low. On these problems, the norms of the minibatch gradients shrink during optimization as progress is made towards a local minima, often reaching near zero towards the end of optimization. This is particularly pronounced on the MNIST and CIFAR-10 datasets, which for a long time were the \emph{de facto} standard for testing optimization methods. 

In more recent work, optimization has focused on larger datasets with fewer training passes (often single-pass), where zero-gradient norms are not expected even near the minima due to inherently noisy labels. On these problems, the magnitude of  $\left\Vert g_{t}\right\Vert$ is dominated by noise and the overall scaling factor of normalization layers. Large Language Model (LLM) training exhibits this behavior for instance. The training behavior of these newer problems is very different from classical MNIST / CIFAR behavior, and much of the early theoretical and empirical work on weight decay is not valid for modern problems.

\section{The Induced Weight Dynamics of Normalized Layers}
\label{sec:dynamics}

The steady-state dynamics of the weights when using weight decay have been previously studied by \citet{van2017l2}, which we briefly recap. Firstly, we focus on a single layer's weight in the network, under the assumption that this layer is immediately followed by a normalization operation (LayerNorm or BatchNorm) in the network. We denote the weights for this layer with $x$ and its gradient $g$. When we need to distinguish individual layers numbered $l=0$ to $L$, we use the notation $x_{t,l}$ for the value of the weights $x$ at time $t$ for a layer $l$.

When using the SGD step, the update at timestep $t$ is:
\[
    x_{t+1} = x_{t} - \gamma g_{t} - \gamma \lambda x_{t},
\]
we see that the squared norm of the weights is updated as follows:
\begin{align*}
\left\Vert x_{t+1}\right\Vert ^{2} & =\left\Vert x_{t}-\gamma g_{t}-\lambda\gamma x_{t}\right\Vert ^{2} \notag\\
 & =\left(1-\lambda\gamma\right)^{2}\left\Vert x_{t}\right\Vert ^{2}-2\left(1-\gamma\lambda\right)\gamma\left\langle g_{t},x_{t}\right\rangle +\gamma^{2}\left\Vert g_{t}\right\Vert ^{2}.
\end{align*}
Here $\langle\cdot,\cdot\rangle$ denotes the inner product. Normalized layers have a key property that allows us to analyze their training dynamics in a simplified fashion. \emph{Their gradients are orthogonal to their weights}, i.e. $\left\langle g_{t},x_{t}\right\rangle =0$. Therefore:
\begin{align}
\left\Vert x_{t+1}\right\Vert ^{2}
 & =\left(1-\lambda\gamma\right)^{2}\left\Vert x_{t}\right\Vert ^{2}+\gamma^{2}\left\Vert g_{t}\right\Vert ^{2}. \label{eq:basic-update}
\end{align}
We can solve for the steady state dynamics by considering the behavior when $\left\Vert x_{t+1}\right\Vert=\left\Vert x_{t}\right\Vert$:
\begin{align*}
\left\Vert x_{t}\right\Vert ^{2} & =\left(1-\lambda\gamma\right)^{2}\left\Vert x_{t}\right\Vert ^{2}+\gamma^{2}\left\Vert g_{t}\right\Vert ^{2}\\
 & =\left(1-2\lambda\gamma+\lambda^{2}\gamma^{2}\right)\left\Vert x_{t}\right\Vert ^{2}+\gamma^{2}\left\Vert g_{t}\right\Vert ^{2},
\end{align*}
\[
\therefore\left(2\lambda\gamma-\lambda^{2}\gamma^{2}\right)\left\Vert x_{t}\right\Vert ^{2}=\gamma^{2}\left\Vert g_{t}\right\Vert ^{2}.
\]
The second-order term in $\lambda$ is extremely small for values of
$\lambda$ and $\gamma$ used in practice,
and so we simplify by dropping this term, to give:
\[
2\lambda\gamma\left\Vert x_{t}\right\Vert ^{2}=\gamma^{2}\left\Vert g_{t}\right\Vert ^{2}
\]
\[
\therefore\frac{\left\Vert g_{t}\right\Vert }{\left\Vert x_{t}\right\Vert }=\sqrt{\frac{2\lambda}{\gamma}}.
\]
\citet{van2017l2} further simplifies with $\left\Vert g_{t}\right\Vert ^{2} \propto \sigma_{\nabla}^{2}/\left\Vert w_{t}\right\Vert ^{2}$, concluding that $\left\Vert x_{t}\right\Vert =\mathcal{O}\left(\left(\gamma/\lambda\right)^{1/4}\right)$. We take a different approach, instead considering the effect of weight decay as controlling the gradient-to-weight ratio $\left\Vert g_{t}\right\Vert / \left\Vert x_{t}\right\Vert$, as it plays a central role in controlling the optimization dynamics.

This argument by \citet{van2017l2} relies on the assumption of an existence of a steady-state in what is a potentially noisy and non-stationary dynamical system. Nevertheless, our experiments indicate that it is highly predictive of the actual behavior of the optimization process. 

\begin{figure}[t]
  \centering
  \includegraphics[width=\linewidth]{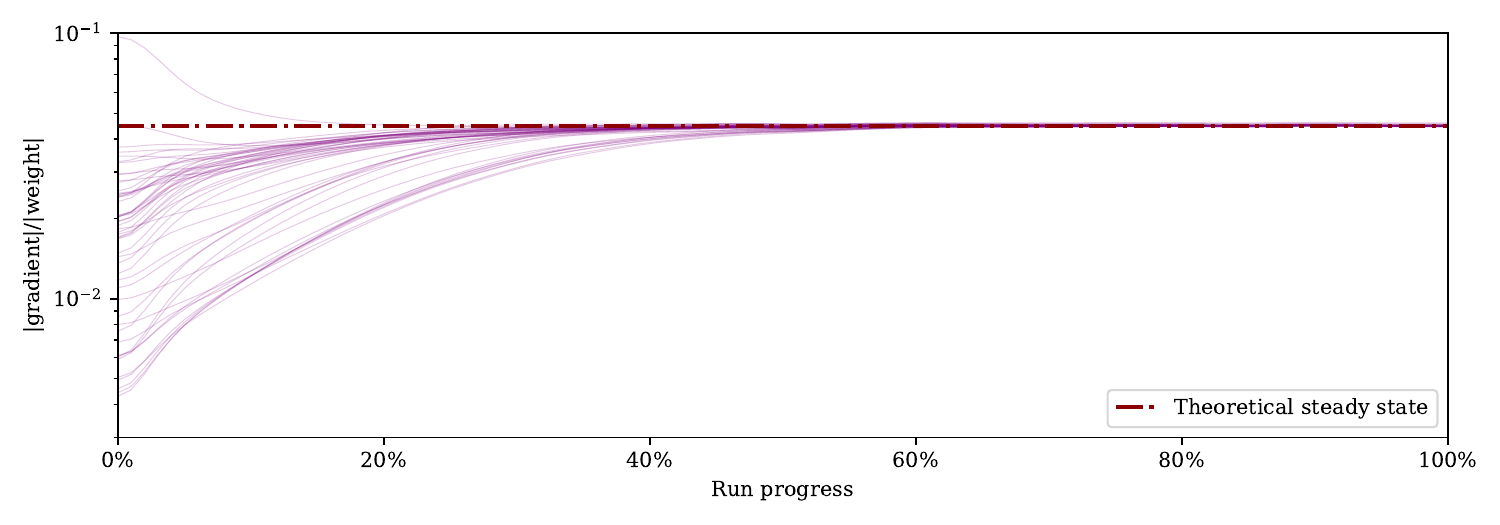}
  \caption{Gradient-to-weight ratios converge towards a steady equilibrium when training without a learning rate schedule. A 100-epoch ImageNet training run using a ResNet-50 model is shown, with each line indicating the ratio for a separate normalized layer. SGD without momentum was used, with LR $0.1$ and weight decay $0.0001$.}
  \label{fig:ratios}
\end{figure}

Figure~\ref{fig:ratios} shows the actual behavior of these gradient-to-weight ratios during training of a ResNet-50 model on the ImageNet dataset when using a flat learning rate with SGD. The value of the ratio predicted by the theory is shown in red. Gradient norms were smoothed with an exponential moving average filter to remove noise. Initially, each layer starts with a different ratio. These ratios quickly converge towards the steady-state predicted by the theory, and continue at that value for the remainder of the optimization process. This balancing also occurs at the neuron level within layers, but for the purposes of understanding global gradient statistics working at the layer-level is sufficient.

\citet{xie2024implicit} also derive a relationship between the infinity norm of the weights and the weight decay constant of $\left\Vert x_{t}\right\Vert _{\infty} =\frac{1}{\lambda}$. This relationship differs from the one we derive as we are considering normalized layers, and we work with a very different set of assumptions.

\subsection{Interaction with schedules and momentum}

\begin{figure}[t]
  \centering
  \includegraphics[width=1.0\linewidth]{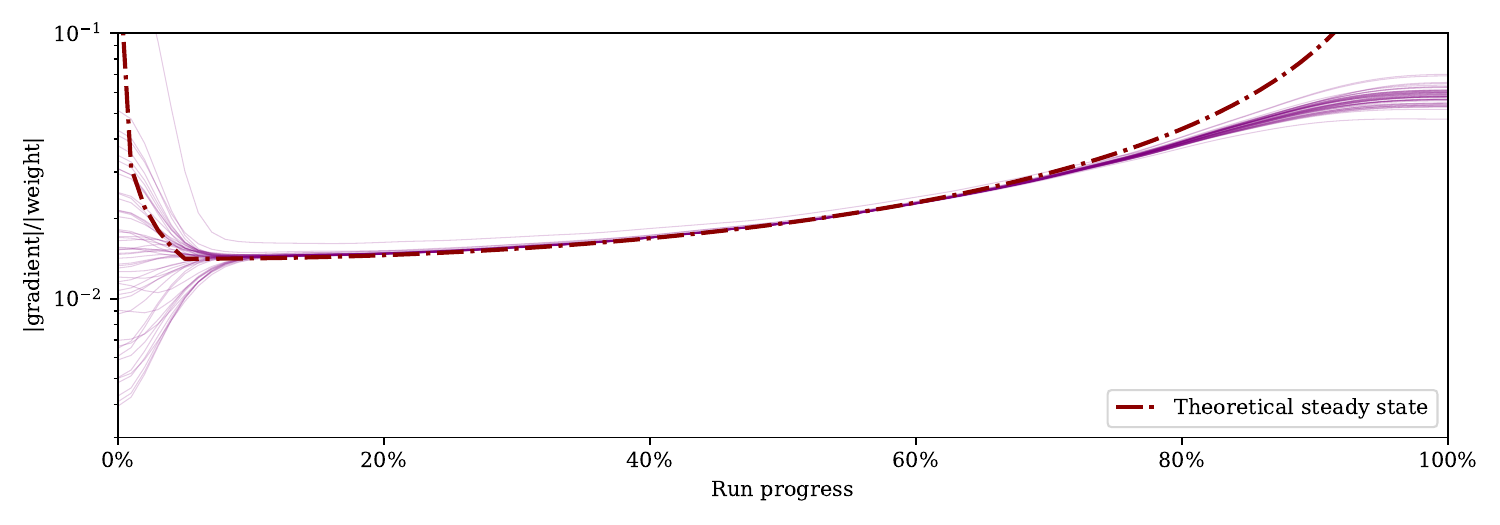}
  \caption{When a cosine learning rate schedule is used, the gradient-to-weight ratios are affected by the schedule. A 100-epoch ImageNet training run using a ResNet-50 model is shown, with each line indicating the ratio for a separate normalized layer. SGD with momentum 0.9, learning rate $0.1$ and weight decay $0.0001$ was used}
  \label{fig:ratios-with-schedule}
\end{figure}

When using a learning rate schedule, the steady-state becomes a moving target, depending on the current learning rate $\gamma_t$. 
\begin{equation}
\frac{\left\Vert g_{t}\right\Vert }{\left\Vert x_{t}\right\Vert }=\sqrt{\frac{2\lambda}{\gamma_t}}. \label{eq:steady-state}
\end{equation}
This behavior can be seen in Figure~\ref{fig:ratios-with-schedule}, where a cosine learning rate schedule with warmup is used. The predicted ratio is shown in dark red. The behavior can be characterized by three stages:
\begin{description}
  \item[Burn-in phase] Where ratios rapidly approach the steady-state value
  \item[Stationary steady-state] Where ratios closely track the steady-state
  \item[Tail blow-up] Near the end, both the steady-state prediction and the actual ratios increase rapidly
\end{description}
The behavior at the end is explained by two factors that counter-act each other; A rapid increase of theoretical steady-state value due to the learning rate in Equation~\ref{eq:steady-state}, which is approaching zero, appearing in the denominator; and the slowing of the optimization process by the decreasing learning rate. This lower learning rate slows the speed at which the ratios can chase the steady-state (Equation~\ref{eq:basic-update}), causing them to no longer track each other closely.

When using SGD with momentum, the steady-state prediction must be modified to account for the fact that the effective learning rate is modified when momentum is used. The effective learning rate with standard SGDM implementations such as in Pytorch and Jax is $\gamma/(1-\beta)$ \citep{defazio2021factorial}, where $\beta$ is the momentum factor. For instance, when using typical values of $\gamma=0.1$ and $\beta=0.9$, the effective learning rate is actually $1.0$. Alternatively, setting the dampening parameter $\tau$ to $\tau=\beta$ keeps the effective step size as $\gamma$. Figure~\ref{fig:ratios-with-schedule} uses SGD with momentum, illustrating that our theory is still highly predictive even in the presence of momentum.

\subsection{Weight decay balances the optimal learning rates of all normalized layers}

Weight decay results in the gradient-to-weight ratios of ALL normalized layers separately converging to the same steady state ratio:
\[
\frac{\left\Vert g_{t,1}\right\Vert }{\left\Vert x_{t,1}\right\Vert }=\frac{\left\Vert g_{t,2}\right\Vert }{\left\Vert x_{t,2}\right\Vert }=\dots=\sqrt{\frac{2\lambda}{\gamma_t}}.
\]
The fact that all normalized layers converge to the same ratio has a critical effect on the optimization process, as this ratio controls, in idealized settings, each layers effective rate of learning.  This means that we can set one global learning rate, without having to tune the learning rates on a per-layer basis. We call this behavior \emph{layer balancing}. 

Balancing the rate-of-learning across layers has been explored in prior works.  \citet{defazioscaling2022} show that for networks with ReLU activations, balancing the gradient-to-weight ratios results in the diagonal blocks of the Hessian having the same average singular value. The Fromage optimizer \citep{bernstein2020distance} explicitly rescales the step for each layer to control the gradient-to-weight ratio. \citet{kosson2024rotationalequilibriumweightdecay} consider effective learning rates as the size of the angular update for normalized layers. Through this framework they also show that a balancing effect is induced by weight-decay.

\section{Weight Decay with AdamW}
\label{sec:adamw}
We can use the same reasoning as we used to analyze the dynamics of
the weights under SGD to consider the AdamW case, however now it requires
a more approximate argument. We will assume no momentum is used, so
that $\hat{m}_{t}=g_{t}$. 

Let us denote Adam's normalized step direction more formally as multiplication
by a diagonal matrix:
\[
A_{t}^{-1}=\text{diag}\left(1/\left(\sqrt{\hat{v}_{t}}+\epsilon\right)\right).
\]
Then the step can be written as:
\[
x_{t+1}=x_{t}-\gamma A_{t}^{-1}g_{t}-\gamma\lambda x_{t}.
\]
Now consider the change in the weights in the matrix-weighted Euclidean
norm $\left\Vert x\right\Vert _{A_{t}}^{2}=x^{T}A_{t}x$:
\begin{align*}
\left\Vert x_{t+1}\right\Vert _{A_{t}}^{2} & =\left\Vert x_{t}-\gamma A_{t}^{-1}g_{t}-\gamma \lambda x_{t}\right\Vert _{A_{t}}^{2}\\
 & =\left(1-\gamma \lambda\right)^{2}\left\Vert x_{t}\right\Vert _{A_{t}}^{2}-2\left(1-\gamma\lambda\right)\gamma\left\langle A_{t}^{-1}g_{t},A_{t}x_{t}\right\rangle 
 + \gamma^{2}\left\Vert A_{t}^{-1}g_{t}\right\Vert _{A_{t}}^{2}\\
 & =\left(1-\gamma \lambda\right)^{2}\left\Vert x_{t}\right\Vert _{A_{t}}^{2}+\gamma^{2}\left\Vert g_{t}\right\Vert _{A_{t}^{-1}}^{2}.
\end{align*}

If we further assume that $A_{t} \approx A_{t+1}$ at the steady state, then
it follows by the same argument as in the SGD case that the steady
state is given by:
\[
\frac{\left\Vert g_{t}\right\Vert _{A_{t}^{-1}}}{\left\Vert x_{t}\right\Vert _{A_{t}}}=\sqrt{\frac{2\lambda}{\gamma}}.
\]

\subsection{AdamW, not Adam, controls norm ratios}
\label{sec:why-adamw}
The argument is not applicable when the weight decay parameter is also normalized by $A^{-1}_t$, as in the original Adam method:
\[
x_{t+1}=x_{t}-\gamma A_{t}^{-1}g_{t}-\gamma\lambda A_{t}^{-1}x_{t},
\]
The change in the norm becomes:
\begin{align*}
\left\Vert x_{t+1}\right\Vert _{A_{t}}^{2} & =\left\Vert x_{t}-\gamma A_{t}^{-1}g_{t}-\lambda A_{t}^{-1}x_{t}\right\Vert _{A_{t}}^{2}\\
 & =\left\Vert x_{t}\right\Vert _{A_{t}}^{2}-\left(2\lambda-\lambda^{2}\right)\left\Vert x_{t}\right\Vert _{A^{-1}_{t}}^{2}+\gamma^{2}\left\Vert g_{t}\right\Vert _{A_{t}^{-1}}^{2}.
\end{align*}
Since the two $x$ norm terms use different weighted norms, the same derivation no longer applies, and the steady-state dynamics for different layers will no longer converge to the same uniform ratio. We conjecture that this lack of balance is a major contributing factor to the performance difference between Adam and AdamW. This conclusion complements the result of \citet{kosson2024rotationalequilibriumweightdecay}, who also consider the differences between Adam and AdamW as arising from unbalanced layer-wise effective learning rates. They consider the size of  angular updates from the two methods, and show that Adam results in unbalanced angular updates when the gradient norm varies between layers.

\subsection{On the learning rate scaling properties of Adam}
In practice the quantity $\left\Vert g_{t}\right\Vert _{A_{t}^{-1}}$
is closely approximated by $\sqrt{\left\Vert g_{t}\right\Vert _{1}}$, as the Adam step approximates a SignSGD
step after an initial burn-in period \citep{bernstein2018signsgd}, since $A_{t}\approx\text{diag}\left(\left
|g_{t}\right|\right)$ (considering the
gradient element-wise with $d$ components):
\[
\left\Vert g_{t}\right\Vert _{A_{t}^{-1}}
=
\sqrt{
\sum_{i}^{d}g_{ti}^{2}A_{tii}^{-1}} \approx 
\sqrt{\sum_{i}^{d}\frac{g_{ti}^{2}}{\left|g_{ti}\right|}}
=\sqrt{\sum_{i}^{d}\left|g_{ti}\right|}
=\sqrt{\left\Vert g_{t}\right\Vert _{1}}.
\]
The weighted norm of $x$  can also be approximated in a non-rigorous fashion by its upper bound from H\"older's inequality: 
\[
\left\Vert x_{t}\right\Vert _{A_{t}}\approx\left\Vert x_{t}\right\Vert _{\infty}\sqrt{\left\Vert g_{t}\right\Vert _{1}},
\]
giving
\[
\frac{\left\Vert g_{t}\right\Vert _{A_{t}^{-1}}}{\left\Vert x_{t}\right\Vert _{A_{t}}}\approx
\frac{\sqrt{\left\Vert g_{t}\right\Vert_{1}}}{\left\Vert x_{t}\right\Vert _{\infty}\sqrt{\left\Vert g_{t}\right\Vert_{1}}}=\frac{1}{\left\Vert x_{t}\right\Vert _{\infty}}.
\]
So AdamW can be interpreted as balancing the layer-wise infinity norms
of the weights:
\[
\left\Vert x_{t}\right\Vert _{\infty} \approx \sqrt{\frac{\gamma}{2\lambda}}
\] The infinity norm of the weights is a key measure
of scaling that determines the correct step size for a layer within
Adagrad family methods (such as Adam and AdamW) \citep{adagrad, orabona2019modern,defazio2023dadapt}, and so balancing this quantity
may allow the same global learning rate to be optimal for every layer when the AdamW optimizer is used.

\section{Correcting Weight Decay}

\begin{algorithm}[t]
    
    
        
    


\begin{algorithmic}[1]
    \State {\bfseries Input:} Initial values $x_{1,l}$ for all layers $l$, 
    \State {\bfseries Input:} $\text{learning rate schedule } \gamma_t, \text{maximum } \gamma_{\max}, \text{decay } \lambda, \beta_1, \beta_2, \epsilon$
    \State $v_{0,l} = m_{0,l} = 0$
    \For{$t=0$ {\bfseries to} T}
    \For{layer $l=0$ {\bfseries to} L}
        \State $g_{t,l} \in \partial f(x_{t,l}, \zeta_{t,l})$ \Comment{Minibatch gradient}
        \State $m_{t,l} = \beta_1 m_{t-1,l} + (1-\beta_1) g_{t,l} $ \Comment{Standard Adam updates}
        \State $v_{t,l} = \beta_2 v_{t-1,l} + (1-\beta_2) \, g_{t,l} \circ g_{t,l}$
        \State $\hat{m}_{t,l} = m_{t,l} / (1-\beta_1^t)$
        \State $\hat{v}_{t,l} = v_{t,l} / (1-\beta_2^t)$
        \If{Layer $l$ is Normalized}
            \State $x_{t,l} = x_{t-1,l} -\gamma_t \hat{m}_{t,l} /(\sqrt{\hat{v}_{t,l}}+\epsilon) - 
        \highlight{\frac{\gamma^2_t}{\gamma_{\max}} \lambda}
         x_{t-1,l}$ \Comment{Corrected weight decay}
        \Else
            \State $x_{t,l} = x_{t-1,l} -\gamma_t \hat{m}_{t,l} /(\sqrt{\hat{v}_{t,l}}+\epsilon) - 
        \highlight{\gamma_t \lambda}
         x_{t-1,l}$ \Comment{Regular AdamW weight decay}
        \EndIf
        
    \EndFor
    \EndFor
    \State Return $x_{T,l}$
\end{algorithmic}
\caption{\label{alg:corrected}Adam with Corrected Weight Decay (AdamC)}
\end{algorithm}

The use of uncoupled weight decay, which a number of works \citep{loshchilov2018decoupled, schaipp2023decaynomore} investigate, is not sufficient to fix tail blow-up behavior above. Consider the uncoupled step, where the learning rate does not multiply the weight decay term:
\[
x_{t+1}=x_{t}-\gamma_t g_{t}-\lambda_t x_{t},
\]
then the gradient-to-weight-ratio steady state becomes:
\[
\frac{\left\Vert g_{t}\right\Vert }{\left\Vert x_{t}\right\Vert }=\frac{\sqrt{2\lambda_t}}{\gamma_t}.
\]
This quantity still depends on the learning rate and so when a learning rate schedule is used, it will again vary over time.

We propose a correction to apply to the weight decay term to completely decouple the steady-state target from the the learning rate schedule. This correction is simply:
\[
\hat{\lambda}_t=\lambda\frac{\gamma_t}{\gamma_{\text{max}}},
\]
where $\gamma_{\max}$ is the maximum learning rate during the schedule, and $\hat{\lambda}$ is the corrected weight decay parameter. Using this value results in a steady state of:
\[
\frac{\left\Vert g_{t}\right\Vert }{\left\Vert x_{t}\right\Vert }=\sqrt{\frac{2\hat{\lambda}_{t}}{\eta}}=\sqrt{\frac{2\lambda\frac{\gamma_t}{\gamma_{\text{max}}}}{\gamma_t}}=\sqrt{\frac{2\lambda}{\gamma_{\text{max}}}}.
\]
The inclusion of $\gamma_{\max}$ here is to keep the scale of the weight decay consistent with the uncorrected case. We call this version of SGD and Adam, with the correction only applied to the normalized layers, SGDC and AdamC respectively. We give pseudo-code for the Adam variant in Algorithm~\ref{alg:corrected}. The weight decay correction is highlighted in yellow.

\section{Experimental Validation}

We focus our experimental validation on two problems: ImageNet training with a ResNet-50 model, and Large-Language Model (LLM) pre-training. These two problems are good exemplars of the phenomenon of increasing gradient norms. Many of the test problems used in the optimization literature for evaluating new optimization methods do not exhibit this phenomenon at the end of training, either because they lack normalized layers, have too short training durations, or suffer from significant overfitting. \citet{defazio2023when} give gradient norm plots for a number of common test bench problems, showing this mixed behavior in practice.

\begin{figure}
\centering \includegraphics[width=\linewidth]{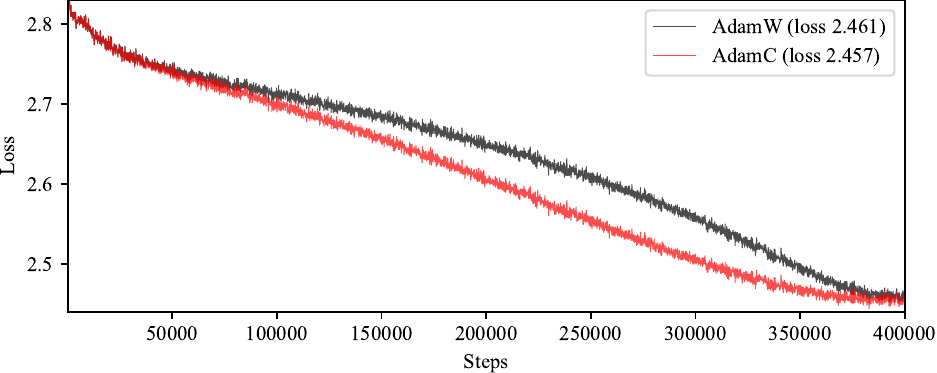}
\newline
\newline
\includegraphics[width=0.5\linewidth]{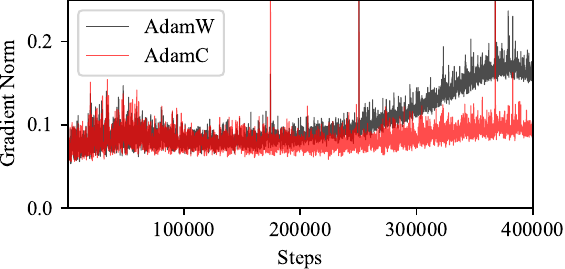}\includegraphics[width=0.5\linewidth]{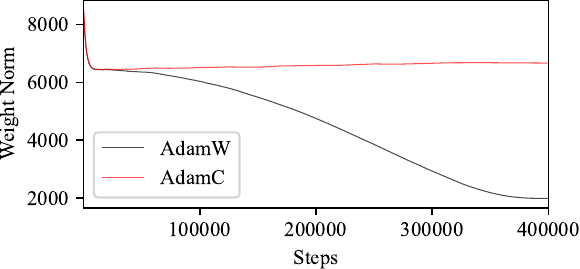}

  \caption{200B token training of a 120M parameter Llama 3 architecture language model trained on FineWeb-Edu}
  \label{fig:llm}
\end{figure}

Figure~\ref{fig:llm} shows the losses, gradient norms and weights norms of a 200B token pretraining run for a small 120M parameter language model using the Llama 3 architecture and trained with a cosine schedule on FineWeb-Edu \citep{penedo2024finewebdatasetsdecantingweb}. The long 200B training duration is important here as a shorter 10B token duration (often used for optimizer evaluation) is too short to show the gradient blowup behavior. 

The application of normalization operations in Transformers no longer corresponds to the direct linear-followed-by-norm structure that our theory covers. For the purposes of applying the correction we consider every linear layer as normalized, excluding the output layer of the network. We trained the model using a 256k token batch-size, weight decay 0.05, and we performed a learning rate sweep on a power-of-two grid for both optimizers separately between $\gamma=0.001$ and $\gamma=0.02$. 

The AdamC correction results in a significant change to the shape of the loss curve, resulting in notably lower loss values. The correction largely removes the increase in the gradient norm during the second half of training. The effect of the correction on the weight norms is particularly clear; AdamW results in rapidly decreasing weight norm whereas AdamC is far more stable.

\begin{figure}
\centering \includegraphics[width=\linewidth]{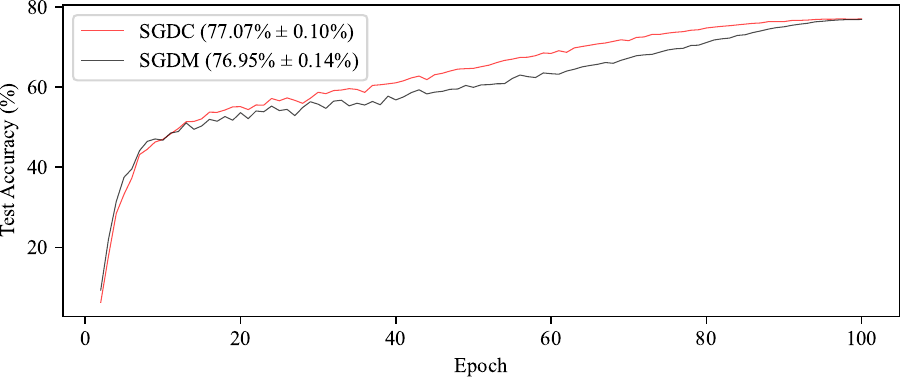}
\newline
\newline
\includegraphics[width=0.48\linewidth]{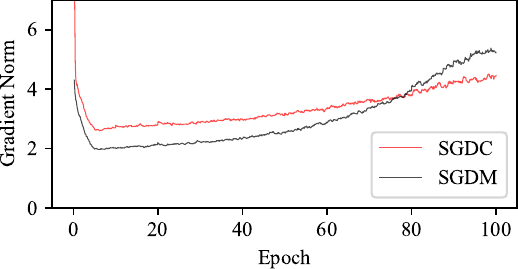}\hfill\includegraphics[width=0.48\linewidth]{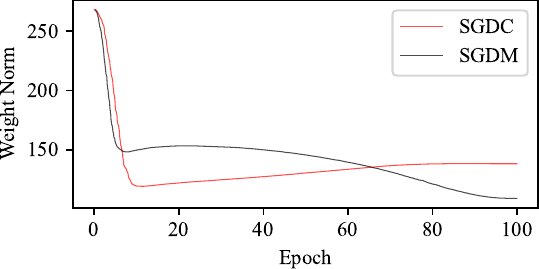}

  \caption{ImageNet ResNet-50 training}
  \label{fig:imagenet}
\end{figure}

Figure~\ref{fig:imagenet} shows the test accuracies, gradient norms and weight norms during training of a standard ResNet-50 model \citep{he2016deep} on the ILSVRC 2012 ImageNet \citep{ILSVRC15} classification dataset using SGD with momentum, together with a cosine learning rate schedule. Following standard practice, we plot the mean of five seeds for both methods. For both methods, we swept the learning rate on a power-of-two grid between $\gamma=0.01$ and $\gamma=1.0$ and the weight decay between $\lambda=0.00005$ and $\lambda=0.01$, and batch-size 256. For ImageNet training, the gradient norm shows an overall linear upward trend over time which is separate from the uptick explained by our theory. When applying SGDC, the rapid increase at the end of training is eliminated, but the slow increase over time remains. 

\section*{Conclusion}
We provide an explanation for a critical part of the behavior of the gradient norm during training. This behavior --- a consequence of the interaction of weight decay and learning rate schedules -- is predicted precisely by the theory we develop, furthermore our theory suggests a simple fix which eliminates the gradient norm blow-up near the end of training. Our theory only addresses one component of the gradient norm behavior, and a number of open questions still remain. Can any remaining drift in the gradient norm be eliminated without using stronger corrections such as projection?

\bibliography{main}
\bibliographystyle{apalike}

\end{document}